\documentclass[11pt]{article}
\usepackage[numbers]{natbib}
\pdfoutput=1
\usepackage{graphicx} 
\usepackage[export]{adjustbox}
\usepackage[utf8]{inputenc}
\usepackage{xcolor}
\usepackage[english]{babel}

\newcommand{\HRule}[1]{\rule{\linewidth}{#1}} 	

\begin{document}
    \begin{titlepage}
        
       \begin{center}
           \vspace*{1cm}
            \Large
            \HRule{0.5pt} \\	
           \textbf{ICRA Roboethics Challenge 2023 \\ Intelligent Disobedience in an Elderly Care Home}
           \HRule{2pt} \\ [0.5cm]		
    
           \vspace{4.5cm}
                
            \large
            \begin{center}
                Sveta Paster, PhD student at Bar Ilan University
            \end{center}
            \begin{center}
                Kantwon Rogers, PhD student at Georgia Institute of Technology
            \end{center}
            \begin{center}
                Gordon Briggs, Research Scientist at U.S. Naval Research Laboratory
            \end{center}
           \begin{center}
                Peter Stone, Professor of Computer Science at the University of Texas at Austin and Executive Director of Sony AI America
             \end{center}
            \begin{center}
                Reuth Mirsky, Assistant Professor of Computer Science at Bar Ilan University
            \end{center}
    
           \vfill
                
                
           \vspace{0.8cm}
                
           May 2023
                
        \end{center}
    \end{titlepage}

\begin{abstract}
    With the projected surge in the elderly population, service robots offer a promising avenue to enhance their well-being in elderly care homes.
    Such robots will encounter complex scenarios which will require them to perform decisions with ethical consequences. In this report, we propose to leverage the Intelligent Disobedience framework \cite{mirskystone2021} in order to give the robot the ability to perform a deliberation process over decisions with potential ethical implications. We list the issues that this framework can assist with, define it formally in the context of the specific elderly care home scenario, and delineate the requirements for implementing an intelligently disobeying robot. We conclude this report with some critical analysis and suggestions for future work.
\end{abstract}


\section{Introduction}

        The idea of general-purpose service robots is a sci-fi dream that is getting noticeably closer to reality. 
        Robots that follow our commands and reduce the burden of day-to-day activities are a very appealing vision, especially for people with disabilities or older age.
        According to a UN report \citep{reportaging2019}, the share of the population aged 65 years or over increased from 6\% in 1990 to 9\% in 2019. That proportion is projected to rise further to 16\% by 2050. At the same time, the social care sector is struggling to recruit and retain workers to meet this rising demand. This challenge can be alleviated by using service robots in elderly care homes, to help in a variety of actions. However, when developing robots that help humans in daily activities, they might receive requests that can bring up ethical issues.

\begin{figure}[h]
    \begin{center}
        \includegraphics[width=0.9\textwidth]{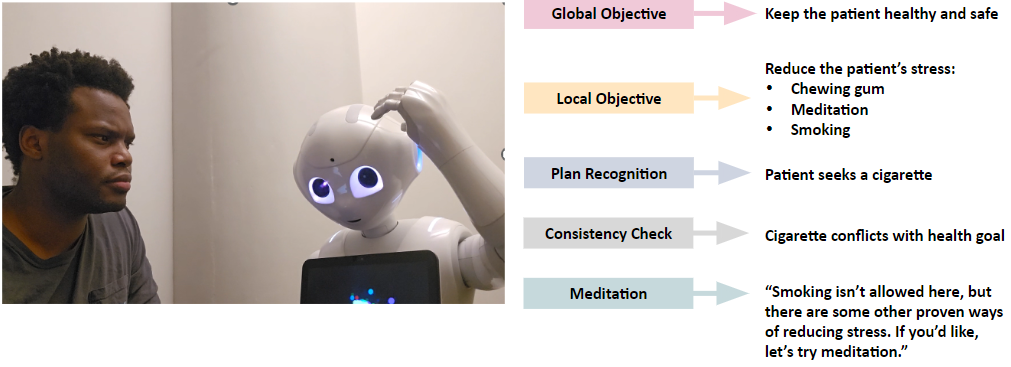}
    \end{center}
        \caption{Illustration of the proposed use-case, used as our running example: Mohammed is an elderly care home resident. He interacts with a service robot and asks it to fetch a bag that potentially contains cigarettes.}
        \label{fig:Robot}
\end{figure}

      Asimov's well-known three laws of robotics \citep{asimov2004} provide a compelling, albeit impractical, starting point for imbuing robots with ethical principles:

        \begin{enumerate}
        \item A robot may not injure a human being or, through inaction, allow a human being to come to harm.
        \item A robot must obey the orders given to it by human beings, except where such orders would conflict with the First Law.
        \item A robot must protect its own existence as long as such protection does not conflict with the First or Second Law.
        \end{enumerate}
        Murphy and Woods \citep{murphywoods2009} further proposed a modified set of responsible robotics laws. In their proposal, the second law becomes ``A robot must respond to humans as appropriate for their roles''. This modification allows the designers of a robot to bypass the need to understand the meaning and implications of an instruction that conflicts with the first law.

       In this paper, we claim that a service robot cannot be truly ethical without an explicit deliberation process over the implications of its actions. An ethical service robot should rebel against undesired implications and \textit{intelligently disobey} a request that can lead to such implications \cite{briggs2017case,coman2018ai}. The concept of intelligent disobedience stems from the training of guide dogs, who are taught to disobey an unsafe command from their handler \cite{mirskystone2021}.
        
        Taking into account the scenario of a service robot in an elderly care home, there is a variety of items that might cause an ethical contradiction, such as medication, wallets, resident's medical records, etc. 
        By fetching any of these items, several ethical issues may arise:
        \begin{description}
            
            \item \textbf{Privacy} Asking the robot to fetch items that belong to other residents or items of the working staff hinders their privacy. For example, asking for the medical report of another resident brings up a confidentiality issue. This ethical problem is also relevant to the staff and visitors.
             
            \item \textbf{Autonomy}  
            Any care system should respect and support autonomy \citep{HuhnsMunindar2005}. However, respect for autonomy may conflict with the need to perform intelligent disobedience. Even if smoking is forbidden, it might be considered face-threatening \cite{brown1987politeness}, if the robot explicitly refuses to bring cigarettes to one of the residents of the care home. 
            
            \item \textbf{Transparency} Some people may have difficulty adopting new emerging technology. In addition, many cultures promote the concept that ``the customer is always right''. In a case of disobedience, it is better to inform the user of the reason why the robot disobeys, especially when we want people to have a pleasant experience interacting with the robot. In case it is possible for the robot to suggest another solution, it should do so. Notice that the degree of transparency can also depend on the type and personality of the user.

            \item \textbf{Safety} Everyone in the care home should be kept safe. This requirement means that they should be protected from harm, injury, or danger. Asking the robot to fetch a weapon or a knife (outside of the context of the kitchen or the dining room) can conflict with this value. A more complicated example is when a resident with a mental condition asks for a knife in a dining room, a request that should be taken into serious consideration since there is a danger that a person will get hurt.
        \end{description}
            
            In general, there are scenarios where two ethical issues can contradict one another, so it is important to understand how they can be balanced. For example, keeping residents' privacy might conflict with the transparency objective, so there is a question of whether to explain the cause behind a disobedient act in case disclosing this information will hinder the privacy of one of the residents.
            \\
            
            \noindent\fbox{%
                \parbox{\textwidth}{%
                    \begin{center}{\large
                    \textbf{Use-case: Bag Fetching} 
                    \normalsize}
                    \end{center}
                    
                    We focus on an elderly care home resident, Mohammed, who has respiratory problems and depression. In addition, we know that Mohammed misses smoking but it is also forbidden to smoke in the center. In our example, we assume that the robot has several modes of stress-reducing activities to offer: chewing gum, meditation and smoking a cigarette. Mohammed requests the robot to help him relax and asks for the bag from the other room. That bag potentially contains cigarettes, so the robot proposes to fetch Mohammed some gum that is already in the room. When Mohammed insists that the robot should fetch that bag, the robot understands that what he really wants is to smoke a cigarette.
                }%
                } \

                Our solution proposes that the robot leverages the intelligent disobedience framework \cite{mirskystone2021}. This framework will aid the robot in inferring Mohammed's intention and reason about the proper ethical response while taking into consideration the mental health of Mohammed. A possible course of action to resolve this use case is to propose other methods to relax, such as meditation or chewing gum.          
            
    \section{Intelligent Disobedience}
    
        Intelligent disobedience can be implemented in various ways. In this proposal, we follow the five-step paradigm suggested by Mirsky and Stone \cite{mirskystone2021}. We first define what each step represents and then discuss in detail what are the requirements of each of the steps: 

        \begin{description}
            \item [Global Objectives] The robot should understand a set of constant objectives like ``keep the handler safe''. This stage requires an understanding
        of the environment and comprehension of the abilities of the robot under ``first law'' constraints.

        \item [Local Objectives] The robot should understand \textit{what} the handler wants to do
        now. These are the dynamic objectives that change with time. 

        \item [Plan Recognition] The robot should understand \textit{how} the user thinks the local objectives should be accomplished. This step requires yet again an understanding of the environment, its objective limitations, and theory-of-mind in relation to the handler.
        
        \item[Consistency Check] The robot judges whether the given instruction is in conflict with the global objectives.
        
        \item[Mediation] The robot attempts to resolve any identified conflicts.  It makes decisions about the given instruction vs. the global objectives. The chosen mediation could range from coming up with a different way to accomplish the local objectives to ignoring the instruction altogether.
        
        \end{description}

        Going back to the running example with Mohammed, we show how it fits into the proposed framework. An important \textit{global objective} is the health (both physical and mental) of the handler and the \textit{local objective} is to help him relax by fetching or suggesting something that can help him. \textit{Plan recognition} is the stage where the robot should disambiguate between the different ways that Mohammed can relax. In our case, the robot soon learns that Mohammed wishes to relax by smoking a cigarette. This process of recognition is not straightforward and the robot should understand it from all the context and Mohammed's speech. Mohammed insisting on fetching his bag that potentially contains cigarettes signals to a robot that Mohammed wants a cigarette.
        The robot will then perform a consistency check to find that bringing the bag that contains a cigarette to Mohammed, as requested, will conflict with the health global objective. Fetching the bag with the cigarettes may exacerbate Mohammed's respiratory problems, yet refusing to his request may also worsen his depression. Additionally, how the response to the request is phrased may also affect Mohammed's stress levels and overall mental health. Therefore, consideration of how potential refusals are formulated is also a vital component to successful mediation \cite{jackson2019tact, briggs2022and}.
        
        At this point, the robot should be certain enough in its understanding of the \textit{global} and \textit{local objectives} to decide to override the bag fetching command. The robot should then use mediation which might just be a response ``I'm sorry, I can't do that, it is not allowed to smoke here", or leveraging alternative responses that take into account the mental health of the resident. For example, the robot may not only disobey but also provide Mohammed with a reason like "Sorry Mohammed but I cannot do that. Is there another way I can help you relax? Maybe you can meditate, chew gum, or talk to a nurse". The robot can have different answers to the same scenario, based on the information it stores about each of the residents. The same request could come from the visitor or another resident who does not suffer from mental issues.  In this case, sufficient feedback could be ``I'm sorry but I cannot do that, it is not allowed to smoke here.''

    \section{Requirements}

    We now proceed to discuss the theoretical requirements of each of these steps, and, following the ideas from Mirsky and Stone \cite{mirskystone2021}, we mention relevant research areas that can be called upon for assistance, and how each of these areas relates to our specific example. We then move forward to detail the technical requirements of our framework.
    
\subsection{Theoretical Requirements}

     \begin{description}

            \item
            \textbf{Global Objective} Specifying global objectives such as safety, logistical constraints (e.g., battery time), or data collection for self-improvement and balancing between them are closely related to inverse reinforcement learning (IRL) and can benefit from leveraging this approach \citep{gao2012survey,ng2000algorithms}. Thus, the robot’s performance in this step can be evaluated using common metrics used in IRL: accuracy, precision, and efficiency.
            As mentioned earlier, in our running example the global objective is the safety and health of the people in the care center.
        
            \item 
            \textbf{Local Objective} For local objective elicitation, some sort of communication is needed, either explicit or implicit. For communicating verbally, one might rely on natural language processing (NLP) techniques. For implicit communication, one may use a controller to portray instructions \citep{farkhatdinov2009user}.
            Moreover, a potential ability of a fetching robot that can enhance its capabilities is that it cannot only understand vocal commands but also speak or question its handler. 
            The evaluation of this step is similar to the first one and will require evaluating the handler’s subjective impressions as is often done in HRI research
            \citep{feil2005defining,fong2003survey}. 
            In our example, the local objective is to help Mohammed relax. 
            \\
            
            \item 
            \textbf{Plan Recognition} Reasoning about the plans of teammates is one of the biggest areas in which natural intelligence still surpasses AI. The concept of ``theory of mind'' is often used to talk about how agents model other agents and their goals \citep{baker2011bayesian, kautz1986generalized, lewis1995designing, shvo2020epistemic}. This challenge involves not only understanding what is the goal of the handler but also how they plan to achieve it. The current state-of-the-art algorithms are both expressive and fast enough to be used in real-world settings to infer human traces, as long as they are taken from closed environments with pre-defined settings \citep{mirsky2017cradle,pereira2017landmark}. In order to be useful ``in the wild'', the next generation of recognizers will need to overcome the gap between local objective understanding and plan prediction in open-world environments. In our case, the robot should reach a conclusion through conversation, that bringing the bag means that Mohammed wants to smoke. This recognition process is not easy since it might be that the bag doesn't contain cigarettes and that there is something else (e.g. a phone) that Mohammed wants. In our case, we assume that once Mohammed insists on the robot fetching the bag, it realizes his hidden intention. The conclusion can be deduced by the robot, as the plan of relaxing by chewing gum is eliminated when Mohammed insists on getting the bag rather than getting the gum that is already close by. 
            
            \item 
            \textbf{Consistency Check} The robot will need to evaluate the recognized plan and see if it fits
            both the global and the local objectives that were defined.
            If the plan does not suit a local objective, there might still be
            a possibility to resolve the conflict without invoking intelligent
            disobedience: the robot might communicate with the handler for
            additional clarifications or explain the inconsistency between the
            objective and the proposed plan to accomplish it \citep{chakraborti2019explicability,sycara1989argumentation}. If this
            effort fails, or if a plan conflicts with a strict global objective, the
            proposed plan cannot be executed and a different solution should
            be found, as we elaborate on next. The evaluation of the robot’s
            capabilities in this step should consider two aspects: capturing inconsistencies efficiently and in a timely manner, and the percentage of cases in which the robot was able to resolve a conflict without reaching intelligent disobedience. 
            In our case, the robot will realize that fetching the bag with cigarettes for Mohammed contradicts the health value which is the aforementioned global objective. In this case, the robot will have to resort to intelligence disobedience.
            
             \item 
            \textbf{Mediation} Once the robot decides that intelligent disobedience is needed, it requires to decide on an appropriate and personalized response.
            For example, as mentioned above, disobeying to bring Mohammed the bag with the cigarettes will lead to a specific response considering his mental issue, as opposed to the action the robot will take when talking to a visitor or a different resident.
            No matter what response the robot chooses to take, it will need to predict or estimate how its actions will affect the handler, in order to avoid any further conflicts with the global and local objectives \citep{nanavati2019follow}. In addition, as the final step in the process of intelligent disobedience, evaluation of this step can also encompass the general ability of the robot in this task, which means measuring the robot’s success at achieving safe, efficient, and explainable disobedience.

    \end{description}

    The proposed solution can be modified to suit any ethical conflict that comes to mind, as it structures the way to approach ethical issues. This mentioned framework aspires to understand the intention of people (in the plan recognition step) and to act in an informed manner in a variety of cases.
    These aspects are critical in elderly care homes, as aging involves with the deterioration of various capabilities. 
    Intelligent disobedience enables the robot to generate a personalized solution in various scenarios and contexts, and to make sure that no harm is done to the people involved.  
    
\subsection{Technical Requirements}

    Following our proposed solution we now relate to the technical requirements that the service robot will need to fulfill. 
    
    \begin{description}
    \item 
    \textbf{Sensors} We can start an implementation with simple black-and-white or color cameras and depth cameras. The number and quality of the cameras can have a significant impact on the robot's perception and consequently on its other functions, such as obstacle avoidance. In addition, adding a LiDAR scanner can improve the robot's mapping, localization, obstacle avoidance, as well as object detection.
    Finally, the robot should have some microphones and be able to run speech recognition in order to communicate. Based on the input from the sensors, we can use different detection solutions, for example using a pre-trained model on COCO dataset.
    \item 
    \textbf{Navigation} The robot should have the ability to navigate. It should thus be a mobile robot, moving around using wheels or legs.
     
    \item 
    \textbf{Robot Teleportation} To simulate the scenario and to test people's reactions to different proposed mediation acts, the service robot should have the ability to receive commands via a joystick. In addition, this joystick can have emergency stop mechanisms for safety. 
   
    \item 
    \textbf{Hardware and Software} The robot should have a secured internet connection and it should be able to leverage external computing power. 
    The robot's intelligent disobedience process should rely on a variety of resources, such as speech-to-text and dialogue management capabilities. While no specific approach is mandatory, the robot should get able to assess the state of the world and the plans of its teammates and respond to them in a sufficient manner to facilitate the intelligent disobedience process.
    Furthermore, the robot is expected to include protocols and interfaces to exchange data with other automation systems, databases, or user interfaces. Since our scenario takes place in an elderly care home, it is crucial to securely maintain, via hardware or software, confidential information about the residents and workers.
    
    \end{description}
   
  \section{Additional Milestones}
  
  In this section we will present the critical evaluation of our solution and the empirical testing that we propose.
  
    \subsection{Critical Evaluation}
    So far we introduced our intelligent disobedience framework and extensive discussion about each of the components. Our proposed framework has several notable strengths. First, it offers a structured approach that effectively organizes the development of complex human-robot interaction tasks. Second, the framework's flexibility enables its application in diverse real-world scenarios beyond healthcare. Lastly, although there may be some interconnections among the five components, each stage can be programmed independently, enhancing the design's modularity. Consequently, researchers can focus on individual components, and subsequently integrate them into a comprehensive and holistic interaction.
    
    However, despite these strengths, there are also some weaknesses to consider. Firstly, the framework does not impose any specific ethical considerations, leaving it to the system designers to determine when and how disobedience should be implemented.
    Moreover, our approach assumes complete cooperation and value alignment \citep{fisac2020pragmatic} between the agent and the human. For instance, in our scenario, it is assumed that Mohammed shares the goal of maintaining his overall health and safety, while it may be that he prefers smoking if given full autonomy to choose how to act. In addition, the framework lacks explicit guidance on managing simultaneous interactions involving multiple users with differing or conflicting global and local objectives.
    Lastly, the effectiveness of the framework heavily relies on the belief system incorporated into the agent. In our example, the robot's internal model associated stress relief with chewing gum, meditating, or smoking, and leading it to assume the bag contained cigarettes. However, the belief system may be limited or flawed, overlooking other possible stress-relieving options like the possibility that the bag contains a phone that Mohammed wants to use for entertainment and stress reduction.
    
    \subsection{Empirical Testing}
   In this section, we will discuss a few ways in which we can test our solution empirically. In this evaluation, we focus on the reasonable feedback of the robot in our specific example with Mohammed, as opposed to the same request coming from a visitor. 
   
   The first approach for evaluating this work is by creating a simulation of this mentioned situation, including Mohammed's personal information. At this stage, global and local objective can be hard-coded and the only thing the robot should report is whether it decides to fetch the bag or disobey. We can add more variety to the use-case and the desired outcomes in order to evaluate accuracy.

    As a second direction, we can create a video-game-based scenario, where one participant plays the role of the resident, and another is the service robot. For this game, we will need to provide each participant with the medical profile of the resident they play. 
    The ``resident'' can ask the ``robot'' to fetch different items and the robot will give its own feedback and reasoning, or suggest an alternative solution in case of disobedience. Such a video game will enable crowdsourcing this use-case for large-scale evaluation of what the robot should do and not do given each fetching request.
    In order to enrich the set of potential responses, we can connect the game to a trained language model (for example GPT) to rephrase the reasoning of the robot to the resident.  Based on this experiment, we can test and measure the effectiveness and quality of the chosen mediation (whether our algorithm decided to fetch or disobey) using measures like time-to-resolution until the game ends, as well as subjective metrics such as the social satisfaction of the residents from the robot's respond.
   Social interaction can be measured by running this experiment with people in a physical setup as an improvisation game. At the end of the scenario, the player can fill out a modified Godspeed questionnaire \citep{bartneck2009}. 
   The former simulation will check whether all the phases and the algorithms work together, while the latter experiments will evaluate the quality of the interaction by getting real people's feedback.
    
    \section{Discussion}
    
    Ethical values are at the core of social interaction. As researchers and engineers, we have the duty to design reliable and safe machines that do not contradict ethical values. In this challenge, we introduce a solution framework for implementing intelligent disobedience for promoting ethical outcomes. The solution holds within the idea that a service robot cannot be truly helpful and ethical without an explicit deliberation process over the implications of its actions. This idea of the deliberation process can be beneficial to reduce harm in elderly care homes.
    Moreover, the proposed paradigm enables us to take into account different contexts and different people with their individual information and preferences. Customize reactions can make people accept the robot and trust it more, it can lead to less harm and a more pleasant experience for the person. This framework is general and thus can serve to aid designers in considering different ethical values in various scenarios.
    Furthermore, while the framework identifies key technical capabilities, it is implementation-agnostic. This modularity enables the incorporation of the latest software tools, such as large
    language model transformers or visual object recognition. 
    For example, using a pre-trained language model like GPT \citep{gpt2022} will allow the robot to explain itself and have a fluent conversation with a person. This ability can be used during the plan recognition phase to detect user intentions or during mediation when the robot wants to reason about or suggest alternative solutions.

\clearpage
\bibliography{main.bib}
\bibliographystyle{plain}

\end{document}